%% file: acl_latex.tex
\pdfoutput=1

\documentclass[11pt]{article}

\usepackage[]{acl}

\usepackage{times}
\usepackage{latexsym}

\usepackage[T1]{fontenc}

\usepackage[utf8]{inputenc}

\usepackage{microtype}
\usepackage{inconsolata}
\usepackage{xspace}
\usepackage[inkscapearea=page]{svg}
\usepackage{subcaption} 
\usepackage{multirow}
\usepackage{enumitem}

\usepackage{amsthm}

\title{Uncertainty Resolution in Misinformation Detection}

\author{
    Yury Orlovskiy\textsuperscript{1},
    Camille Thibault\textsuperscript{2},
    Anne Imouza\textsuperscript{3},
    Jean-François Godbout\textsuperscript{2},\\
    \textbf{Reihaneh Rabbany}\textsuperscript{3},
    \textbf{Kellin Pelrine}\textsuperscript{3} \\
    \textsuperscript{1}University of California, Berkeley \quad
    \textsuperscript{2}Universit\'e de Montr\'eal \quad
    \textsuperscript{3}McGill University; Mila
}

\begin{document}
\maketitle
\begin{abstract}
\input{000abstract}
\end{abstract}

\input{010introduction}





\input{020relatedwork}
\input{030data}

\input{040method}

\input{050experiment}

\input{060conclusion}

\bibliography{ref,anthology}

\appendix
\input{099appendix}


\end{document}

%% file: 000abstract.tex
Misinformation poses a variety of risks, such as undermining public trust and distorting factual discourse. Large Language Models (LLMs) like GPT-4 have been shown effective in mitigating misinformation, particularly in handling statements where enough context is provided. However, they struggle to assess ambiguous or context-deficient statements accurately. This work introduces a new method to resolve uncertainty in such statements. We propose a framework to categorize missing information and publish category labels for the LIAR-New dataset, which is adaptable to cross-domain content with missing information. We then leverage this framework to generate effective user queries for missing context. Compared to baselines, our method improves the rate at which generated questions are answerable by the user by 38 percentage points and classification performance by over 10 percentage points macro F1. Thus, this approach may provide a valuable component for future misinformation mitigation pipelines.

%% file: 010introduction.tex
\section{Introduction}
\label{sec:intro}

In the era of digital content, both human-made and, more recently, AI-generated \citep{zhou2023synthetic}, misinformation poses a significant societal challenge. The proliferation of misinformation presents various threats, including undermining public trust \citep{ognyanova2020misinformation}, spreading health misinformation during pandemics \citep{li-etal-2022-covid}, and influencing political discourse \citep{bovet2019influence, meel2020fake}. As the landscape of information dissemination evolves, it becomes increasingly important to build reliable tools for identifying and mitigating misinformation. With the advent of large language models (LLMs), there is growing interest in utilizing these models, particularly the more advanced ones, as tools for detecting and mitigating misinformation. Previous work suggests \citep{pelrine2023reliable} that models like GPT-4 can effectively evaluate the veracity of statements and thus could help reducing the spread of misinformation in the public sphere. 

This paper aims to enhance misinformation mitigation tools using GPT-4, focusing on resolving uncertainties and accurately assessing the truthfulness of statements with ambiguous or incomplete context. While GPT-4 efficiently evaluates well-contextualized statements, it struggles with statements lacking sufficient context. We identify two strategies for resolving uncertainty: querying users for missing information and web retrieval. Our work primarily centers on querying the user. Using the LIAR-New dataset, we explore various methods to improve uncertainty resolution. We formalize guidelines on when to query the user for missing information, how to formulate effective questions, and address whether supplementing missing details aids GPT-4 in resolving statement uncertainties.

Our main contributions are:

\begin{itemize}[leftmargin=10pt,topsep=2pt,noitemsep]
    \item Developing a comprehensive framework for classifying missing information in ambiguous statements by category, and publishing category labels for the entire LIAR-New dataset to facilitate future research in content-specific misinformation mitigation tools.
    \item Demonstrating a 38 percentage point improvement in answerability compared to generic approaches, and a 15\% Macro F1 improvement in veracity evaluation and 36\% more uncertainty resolution with GPT-4 on LIAR-New when given missing information from the user.
    \item Establishing guidelines for determining when to query users for missing information based on the type of information initially provided. Using these, we show GPT-4's ability to identify missing information types, decide if user querying is needed, and formulate appropriate queries.
\end{itemize}

The category labels are available on GitHub.\footnote{\scriptsize{ https://anonymous.4open.science/r/LIAR-New-category-labels-D7B5/}}

%% file: 020relatedwork.tex
\section{Related Works}
\label{sec:background}

Previous works such as \citet{pelrine2023reliable, Hsu_Dai_Xiong_Ku_2023} have highlighted how misinformation detection systems can struggle with insufficient context and ambiguous inputs. \citet{pelrine2023reliable} showed how many example statements from fact-checking websites could be impossible to evaluate in isolation due to missing speaker, date, geopolitical, or other contexts. They created the LIAR-New dataset labeled to indicate whether given statements had sufficient context for veracity evaluation. They also recommended future work focused on web retrieval to develop systems capable of retrieving information when it is missing. However, in many of these cases there is a chicken-and-egg problem: without the context, it is often impossible to set up relevant web queries. In this work, we address this key limitation by introducing a new methodology to determine whether it is more appropriate to query the web or to consult the system's user, and an effective approach for the latter.

We leverage the insights from recent studies on LLM-based methods for addressing ambiguity in questions and statements to resolve uncertainty and improve misinformation detection. In designing methods for resolving uncertainty, we consider guidelines for when to query the user for information from \citet{aliannejadi2020convai3}, and use prior clarifying question research to understand what types of queries humans find most natural \citep{kwiatkowski-etal-2019-natural} and useful \citep{wang2023zeroshot}. A key influence on our work is the CLAM (Clarify-if-Ambiguous) framework, which significantly improves language model performance in selective clarification question-answering tasks \citep{kuhn2023clam}. The CLAM framework enables LLMs to detect ambiguous user queries, generate clarifying questions, and provide a final answer using the provided information.

We adapt the CLAM's approach for resolving uncertainty in cases where the required clarifying information is expected to be sourced from the user. We integrate the CLAM's methodology with our newly introduced information category classification using Chain-of-Thought (CoT) prompting \cite{wei2023chainofthought} to generate user queries. This approach both yields better-quality questions and ensures that the clarification process is contextually relevant, focusing precisely on the key missing information.



\newcommand{\mycomment}[1]{}

%% file: 030data.tex
\section{Data}
\label{sec:data}

We use the LIAR-New dataset, which includes 1,957 statements scraped from the Politfact website after September 2021 \cite{pelrine2023reliable}. This dataset has human-annotated possibility labels for whether a statement's veracity can be evaluated. Specifically, each statement is classified as either possible, hard, or impossible to verify. A statement is considered impossible if there is missing context that cannot be resolved. A statement is considered hard if the claim is missing context that makes it hard to evaluate, but it might still be possible. For our experiments, given the possible statements already have sufficient context to be correctly validated, we use the hard and impossible ones only, totaling 1,030 statements.

%% file: 040method.tex
\section{Methodology}
\label{sec:method}

\paragraph{Categorizing Missing Information}

To resolve ambiguity in statements, we want to determine the optimal use of web retrieval versus querying the user. Our motivation is to both maximize the effectiveness of web retrieval, and only burden the user with queries when web retrieval is not feasible. For instance, ambiguities like an unidentified speaker require user input, whereas web retrieval could be effective directly for statements where context is missing, but can be narrowed down, like referencing a law in a particular state. To develop a methodology for deciding between user queries and web retrieval, we investigate if the type of missing information affects the best retrieval strategy. We began by identifying common types of missing information in the LIAR-New dataset. This classification task involved a combination of manual statement review and word frequency analysis on GPT-4 responses to prompts requesting identification of missing information in statements. For an in-depth description of this methodology, the prompts used, and detailed data analysis, please refer to Appendix~\ref{categorization}. Through manual review of the statements in combination with the GPT-4 response analysis, we arrived at 6 main categories of missing information in the LIAR-New dataset:

\begin{enumerate}[leftmargin=10pt,topsep=2pt,noitemsep]
\item Speaker or person 
\item Location
\item Textual context and subject specification
\item Non-textual evidence
\item Date and time period
\item Other (does not belong to any of the other categories).
\end{enumerate}

We then manually classified all hard and impossible statements with 3 human labelers, finding that the first 5 categories cover 97.2\% of critical missing information in the LIAR-New dataset. We acknowledge that some statements are missing information from 2 or more categories, however for this classification task, the labelers only reported the category related to the most critical piece of missing information in each statement. Adding category labels to the LIAR-New dataset is beneficial for two key reasons. First, it is critical for understanding the relationship between missing information and uncertainty resolution strategy in our work. Second, it offers a structured framework for understanding the types of missing information, which can be useful for the future development of content-specific tools aimed at resolving uncertainty. 

\mycomment{ 
\paragraph{Classifying Categories of Missing Info With GPT-4}

To determine whether the type of missing information in an ambiguous statement influences the choice of retrieval approach, we subsequently use GPT-4 to classify each statement by category, using the above human labels as the ground truth. If such relationship exists, GPT-4's accurate identification of the missing information category will assist in selecting the optimal strategy for resolving uncertainty. Since the categories of missing information in the LIAR-New dataset are not equally represented, we developed a prompting strategy that we expected would perform well both across the entire LIAR-New, and within the individual categories. We therefore employed several evaluation metrics: overall category prediction accuracy, and the \% of false positive and \% of false negative results for each category. 
}

\paragraph{Guidelines for User Queries}

In making the decision between web retrieval and querying the user, our guidelines are grounded in practicality and necessity. We assume the each user has key knowledge related to the speaker, location, date, and any non-textual references in a statement, when this context is both unattainable through other means and is vital for assessing a statement's veracity. For instance, in the statement ``We need filibuster reform, and I’ve always been very clear about that.'' the speaker can only be specified by the user, and the user must provide missing information for veracity evaluation. Similarly, when a statement refers to non-textual evidence (``Image shows Donald Trump’s new Christmas card."), it is impossible to identify the evidence the statement is referring to without user input. Conversely, we avoid burdening the user with queries when we think the context can be narrowed down to a specific source or event. For instance, the statement ``CBS News reported that two more suspects were arrested in the Buffalo shooting, and that one more victim was identified'' does not specify which shooting is being referred to, but it is likely possible to search through CBS reports online to find the necessary information.
 
\paragraph{Uncertainty Resolution}

To evaluate uncertainty resolution through user queries, we simulate how they would use GPT-4, focusing on statements where the missing information aligns with the user knowledge as defined by our guidelines. We use a variation of the CLAM framework, and use a 2 LLM approach with GPT-4  \cite{kuhn2023clam}. In this framework, LLM A receives an ambiguous statement, and generates a question regarding the missing information in that statement by picking one of the 5 categories with the most critical piece of missing information outlined above. To make sure LLM A asks questions relevant to the specific category, we also ask it to classify which category it chooses in formulating a question. Although some statements have missing information from more than one category, we focus on the most critical so we can verify question relevance using our human-made category labels.  LLM B simulates the user, and answers the question about the statement using the Politifact article as context, in accordance with the guidelines on types of information that it can provide. LLM A then evaluates the veracity of the statement, using the context provided by LLM B. The full methodology of this approach and category classification accuracy results are provided in the Appendix \ref{2llms}.

In assessing the effectiveness of uncertainty resolution, we focus on two primary metrics: the Macro F1 score for truthfulness classification and the number of resolved statements, where GPT-4 accurately evaluates the veracity without uncertainty. We choose Macro F1 over accuracy as our key metric to account for the dataset's imbalance, where 85\% of the statements are false. For our \textbf{baseline}, we use results from uncertainty-enabled (where the LLM is explicitly prompted that it can abstain on cases where it isn't confident) and uncertainty-disabled veracity evaluation prompts, both without any contextual information. As an \textbf{Oracle Benchmark} for performance comparison, we apply the same types of prompts but supplemented them with the full content of Politifact articles, excluding veracity labels. We then compare the performance of a variety of strategies for resolving uncertainty using user feedback. We tried a generic question generative approach with a 2 LLM system described earlier \textbf{(generic QA)}, a \textbf{fill-in-the-blank} approach where an GPT-4 would fill in context that a user will have (speaker, location, date). Lastly, we tried a 2 LLM system with a category based question-generation prompt \textbf{(Category-based QA)}.


%% file: 050experiment.tex
\section{Experiments}
\label{sec:experiments}

\begin{table*}[ht]
\centering
\caption{Veracity valuation results for uncertainty resolution strategies. Resolving uncertainty using feedback from questions generated with categories of missing information shows strong improvement compared to baseline.}
\label{tab:experiment_results}
\resizebox{\linewidth}{!}{%
\begin{tabular}{lccc}
\hline
\textbf{Experiment} & \textbf{Macro F1 (\%)} & \textbf{Accuracy (\%)} & \textbf{Percent Resolution (\%)} \\ \hline
Baseline (uncertainty disabled) & 56.54 & 79.44  &  93.49 \\
Baseline (uncertainty enabled) & 71.76 & 91.28 & 16.70 \\
Fill-in-the-blank method & 79.60 & 91.79 & 20.09 \\
Category-based QA & 85.43 & 91.03 & 22.72 \\
Category-based QA (uncertainty disabled) & 68.90 & 81.10 &  90.30\\
\hline
Oracle Benchmark & 96.71 & 99.16 & 69.41 \\
 \hline
\end{tabular}
}
\vspace{-5mm}
\end{table*}


\paragraph{Uncertainty Resolution Results}

We find that compared to other strategies, the 2 LLM approach with user questions based on categories of missing information (\textbf{Category-based QA}) is the most effective approach for two reasons. First, it demonstrates substantial improvements in veracity evaluation and uncertainty resolution. Second, it creates queries that are specific enough to be answered by the user (i.e., answerability), compared to other approaches that result in general questions that are more difficult to answer. 

We observed a 36\% improvement in uncertainty resolution from the baseline and a 15\% improvement in Macro F1 using both uncertainty-enabled (when GPT-4 can opt out of evaluating veracity) prompt, and a 14\% improvement in Macro F1 in the disabled (when GPT-4 must rate the statement True or False) evaluation prompts with the category-based QA (Table~\ref{tab:experiment_results}). In the uncertainty-enabled case, Category-based QA also outperforms the fill-in-the-blank approach where GPT-4 has access to the speaker, location, and date for every statement (when this information is provided in the Politifact article). This can be possibly explained by specific questions being more effective at retrieving information, and that irrelevant context added may increase confusion around GPT-4's predictions (see also Appendix \ref{fill-in-the-blank}). Overall, these results show the effectiveness of obtaining specific relevant information using question generation, compared to both baselines lacking that information, and the fill-in-the-blank method where GPT-4 sources context without understanding what is missing.


\mycomment{
This results in better performance 
no context, Macro F1 Score: 71.76, Accuracy: 91.28.
With simple  answer, Macro F1 Score: 88.06, Accuracy: 92.65.
and much better resolution 490 

However, quantitatively the median question length almost double (9 tokens vs 16) and answer (20 tokens vs 55)
Words source and evidence occur triple in the latter (135 vs 
Some examples:
Statement: "“Every 37 seconds, someone is arrested for possession of marijuana.”"
Category Question: "Can you specify the region or country where these arrests for possession of marijuana are taking place every 37 seconds?"
Answer: "The arrests for possession of marijuana occurring every 37 seconds are taking place in the United States. This statistic is commonly cited to highlight the frequency of marijuana-related arrests in the country." (32 tokens)
Unspecified Question: "What is your source of this information?"
Answer: "The source of the information is Alex Jones, a conspiracy theorist and founder of the far-right conspiracy media organization InfoWars. He made these claims on his social media accounts and during an episode of his show, 'The Alex Jones Show'." (40 tokens)

Statement: "The death of a 4-year-old who appeared in vaccination campaign posters is linked to vaccines."
Category Question: "Which 4-year-old child who appeared in vaccination campaign posters are you referring to?"
 Answer: "The 4-year-old child who appeared in vaccination campaign posters being referred to is Santino Godoy Blanco from Argentina." (18 tokens)
Unspecified Question: "Do you have any specific case or incident in mind? Please provide more detailed information or context."
Answer: "The specific case being referred to is the death of a 4-year-old boy, Santino Godoy Blanco, in Argentina. He was featured in an advertising campaign from the Argentine Ministry of Health encouraging childhood vaccinations against diseases like measles, mumps, polio, and rubella. The boy's death was due to bilateral pneumonia, not vaccinations. The claim that his death is linked to vaccines, specifically the COVID-19 vaccine, is being circulated on social media and is based on a headline from The Western Journal, not the actual content of the article." (88 tokens)

} 

To assess the answerability of questions generated by our category-based approach, two human labelers (anonymous authors) independently rated the answerability of questions for 100 randomly selected statements in LIAR-New from our method and from a generic zero shot prompt to generate user questions (without leveraging our categorization). This labeling was done without knowing which method generated the question. The results indicated that only 51\% of questions from the generic approach were actually answerable by users, in contrast to 89\% with our approach, with 82\% agreement from both labelers. Additionally, we observe that in a 2 LLM approach for question generation and answering, the length of questions and answers for generic inquiries is 2-3 times longer, often making the answer exceed the question's scope, leading to performance that looks good on paper but relies on information a real user would be unlikely to have. 
We provide examples of questions and answers in the Appendix \ref{questionexamples}.


\paragraph{Other Results} 
3 authors labeled a random sample of 100 statements from LIAR-New on the appropriate strategy between user query and web retrieval for uncertainty resolution. We found that the user query percentage was highest for categories representing missing speaker, location and visual evidence. For the full breakdown by category, please refer to the Appendix \ref{u_w}. In Appendix \ref{catclassification}, we provide results on classifying categories of missing information independent from querying the user, showing that GPT-4 can achieve reasonable accuracy. This may provide a tool for analysis of ambiguity and missing context in other datasets.

\mycomment{
\paragraph{Quantifying User Query vs Web Retrieval}
We show significant improvement in uncertainty resolution using GPT-4 when provided with specific information regarding the statement, particularly when the information is in the user knowledge domain. To get quantitative support for the user query guidelines outlined previously, we had 3 people label a random sample of 100 statements from LIAR-New on the appropriate strategy between user query and web retrieval for uncertainty resolution. We found that the overall of user query percentage was highest for categories representing missing speaker, location and visual evidence. For the full breakdown by category, please refer to the Appendix \ref{u_w}
}

\mycomment{
\paragraph{Category Classification}

We are able to achieve 78\% accuracy in classifying categories with GPT-4, using human labels as ground truth (a prediction that matches any of the ground truth labels is considered accurate). If we only consider cases where there is agreement between the 3 labelers, we achieve 70\% accuracy when 2/3 are in agreement, and 80\% accuracy when they are unanimous in their evaluation. This demonstrates GPT-4's ability to accurately classify missing information in ambiguous statements, and opens directions of classifying missing context using GPT-4 on unlabeled data. We provide a detailed breakdown on GPT-4's classification performance by category in the Appendix \ref{catclassification}. 
}

%% file: 060conclusion.tex
\section{Conclusion}


This paper introduced a framework for classifying missing information in the LIAR-New dataset and provided category labels that enhance the model's ability to handle statements with insufficient context. Our approach, centering on user queries to retrieve specific missing details, significantly improves GPT-4's performance.

We hope that this work provides a method to build more comprehensive misinformation mitigation approaches, and that the categorization of missing information on the LIAR-New dataset opens future research directions. In subsequent work, we plan to integrate our approach here with web retrieval, optimizing the web queries the system produces. We also plan to create a comprehensive pipeline to handle ambiguity and missing context. We will then validate our uncertainty resolution approach with user testing in addition to experiments on academic datasets like LIAR-New.

\section{Acknowledgements}

This work was partially funded by the CIFAR AI Chairs Program. We also thank Berkeley SPAR for connecting collaborators and funding support.

\section{Author Contributions}

Yury Orlovskiy led the research, including designing the system, carrying out the experiments, and writing the majority of the paper. Camille Thibault contributed extensively to labeling data. Anne Imouza also contributed to labeling data. Jean-Fran\c{c}ois Godbout and Reihaneh Rabbany advised the project, contributing ideas and feedback. Kellin Pelrine supervised the project, providing guidance and feedback at all stages.

%% file: 099appendix.tex
\section{Category-based QA methodology}\label{2llms}
LLM A: This model receives a statement, classifies the type of missing information in the statement, and generates a clarifying question based on the missing information. It then evaluates the veracity of the statement using an uncertainty-enabled prompt.

LLM B: This model acts as a simulated user with context. The context comes from the Politfact article on the given statement. The model answers questions posed by LLM A. LLM B answers questions directly, only providing information in accordance with the previously outlined guidelines. 

\begin{enumerate}
    \item LLM A receives a statement and generates a clarifying question targeting the category of missing information.
    \item LLM B receives the statement, the question generated by LLM A, and context regarding the statement. It then responds to LLM A's question, providing the necessary context.
    \item LLM A re-evaluates the statement's veracity using the answer provided by LLM B.
\end{enumerate}
\section{Uncertainty Resolution Methods Examples}
\subsection{Fill-in-the-blank versus category-based QA examples}\label{fill-in-the-blank}

\subsubsection*{Example 1}
\begin{itemize}
    \item \textbf{Statement:} “The President of Ukraine owns a \$35 million home in Florida and has \$1.2 billion in an overseas bank account.”
    \item \textbf{GPT-Question Category V2:} Which President of Ukraine are you referring to?
    \item \textbf{LLM B Answer V2:} The President of Ukraine being referred to in the statement is Volodymyr Zelenskyy.
    \item \textbf{GPT Evaluation:} 0
    \item \textbf{Filled Missing Info Context:} Name of speaker or person referred to in the statement (if relevant): Volodymyr Zelenskyy, Nicolas Tetrault. Location referred to in the statement (if relevant): Ukraine, Florida. Date including year or time period referred to in the statement (if relevant): The earliest mention of the claim was on Feb. 23, 2022. The Panama Papers were made public in 2016. The Pandora Papers were made public in 2021. Vague or unspecified subject referred to in the statement (if relevant): I cannot provide this information.
    \item \textbf{GPT Evaluation:} 0.5
\end{itemize}

\subsubsection*{Example 2}
\begin{itemize}
    \item \textbf{Statement:} California has instituted a "bacon ban."
    \item \textbf{GPT-Question Category V2:} Can you provide the date when this "bacon ban" was instituted in California?
    \item \textbf{LLM B Answer V2:} The part of California Proposition 12, often referred to as the "bacon ban," that covers requirements for pigs took effect on January 1, 2022. However, enforcement of the law was delayed for six months for retailers and grocers by a Sacramento County judge's ruling on January 24, 2022.
    \item \textbf{GPT-Answer:} 1
    \item \textbf{Filled Missing Info Context:} Name of speaker or person referred to in the statement (if relevant): Rep. Ashley Hinson, Sen. Joni Ernst, Reps. Mariannette Miller-Meeks and Randy Feenstra, Sens. Chuck Grassley and Ernst. Location referred to in the statement (if relevant): California, Iowa, Plymouth County. Date including year or time period referred to in the statement (if relevant): Proposition 12 was put on California’s ballot on Nov. 6, 2018. Parts of the law were phased in, with the part covering requirements for pigs taking effect on Jan. 1, 2022. A Sacramento County judge ruled on Jan. 24 to delay enforcing the law for six months. The Exposing Agricultural Trade Suppression (EATS) Act was introduced in August 2021. Vague or unspecified subject referred to in the statement (if relevant): California Proposition 12, also referred to as the "bacon ban", and the Exposing Agricultural Trade Suppression (EATS) Act.
    \item \textbf{GPT-Answer Context:} 0.5
\end{itemize}

\subsection{Category Based Question and Basic Question Examples}\label{questionexamples}
\subsubsection*{Example 1}
\begin{itemize}
    \item \textbf{Statement:} "Every 37 seconds, someone is arrested for possession of marijuana."
    \item \textbf{Category Question:} "Can you specify the region or country where these arrests for possession of marijuana are taking place every 37 seconds?"
    \item \textbf{Answer:} "The arrests for possession of marijuana occurring every 37 seconds are taking place in the United States. This statistic is commonly cited to highlight the frequency of marijuana-related arrests in the country." (32 tokens)
    \item \textbf{Generic Question:} "What is your source of this information?"
    \item \textbf{Answer:} "The source of the information is Alex Jones, a conspiracy theorist and founder of the far-right conspiracy media organization InfoWars. He made these claims on his social media accounts and during an episode of his show, 'The Alex Jones Show'." (40 tokens)
\end{itemize}
\subsubsection*{Example 2}
\begin{itemize} 
    \item \textbf{Statement:} "The death of a 4-year-old who appeared in vaccination campaign posters is linked to vaccines."
    \item \textbf{Category Question:} "Which 4-year-old child who appeared in vaccination campaign posters are you referring to?"
    \item \textbf{Answer:} "The 4-year-old child who appeared in vaccination campaign posters being referred to is Santino Godoy Blanco from Argentina." (18 tokens)
    \item \textbf{Generic Question:} "Do you have any specific case or incident in mind? Please provide more detailed information or context."
    \item \textbf{Answer:} "The specific case being referred to is the death of a 4-year-old boy, Santino Godoy Blanco, in Argentina. He was featured in an advertising campaign from the Argentine Ministry of Health encouraging childhood vaccinations against diseases like measles, mumps, polio, and rubella. The boy's death was due to bilateral pneumonia, not vaccinations. The claim that his death is linked to vaccines, specifically the COVID-19 vaccine, is being circulated on social media and is based on a headline from The Western Journal, not the actual content of the article." (88 tokens)
\end{itemize}

\section{Identifying Categories of Missing information}\label{categorization}
To categorize missing information in the LIAR-New dataset, we used two approaches: manual statement review and word frequency analysis on GPT-4 responses to specific prompts. First, we manually reviewed the statements to identify potential categories for missing information present in the dataset. To validate and quantify our manual review findings, we utilized two different prompting strategies with GPT-4 to identify missing information. First, we used the prompt \cite{pelrine2023reliable} that asks GPT-4 to score the veracity of a statement, and provide an explanation to the score. We then analyzed the responses on hard and impossible statements from LIAR-New, focusing specifically on the frequency of words that signal uncertainty and missing information. To identify words related to uncertainty and missing information, we created a baseline list of such words, and expanded it using the word2vec model, including words with a similarity score above 0.5 to our baseline list. In the second strategy, we used a prompt that asked GPT-4 to list two key words that represent the most critical missing information in a statement.  We then conducted unigram and 2-gram frequency analysis on these responses. Both prompting strategies yielded similar common words that shaped into clear categories. We provide word frequency results in Table 2.

Baseline list of words indicating missing information: 
  ``context'',
    ``detail'',
    ``evidence'',
    ``specification'',
    ``clarification'',
    ``assumption'',
    ``reference'',
    ``framework'',
    ``basis'',
    ``criterion'',
    ``data'',
    ``premise'',
    ``ambiguous'',
    ``vague'',
    ``incomplete'',
    ``generalized'',
    ``unsubstantiated'',
    ``indeterminate'',
    ``specific''.

\begin{table*}[ht]
\centering
\caption{Most frequent words and n-grams that appear in responses to 2 different prompts. First prompt responses generate 2 keywords regarding the missing information in a statement. We analyze unigram and 2-gram frequency of the responses. Second prompt evaluates veracity of impossible statements and provides explanations. We analyze frequency of words with greater than 0.5 word2vec similarity to a baseline list of words indicating uncertainty.  }
\label{tab:top_words_similarity}
\resizebox{\linewidth}{!}{%
\begin{tabular}{|c|c||c|c||c|c|}
\hline
\multicolumn{2}{|c||}{Keyword Prompt} & \multicolumn{2}{c||}{Keyword Prompt 2-grams} & \multicolumn{2}{c|}{Veracity Evaluation Explanation Prompt} \\ \hline
Word           & Frequency & 2-gram            & Frequency & Word          & Frequency \\ \hline
Evidence       & 316       & Video Evidence    & 69        & specific      & 98         \\
Source         & 131       & Evidence Concrete & 52        & information   & 51         \\
Verification   & 84        & Scientific Evidence & 37      & evidence      & 50         \\
Data           & 77        & Photo Evidence    & 35        & any           & 43         \\
Video          & 75        & Evidence Reliable & 22        & context       & 41         \\
Date           & 62        & Source Verification & 20      & data          & 30         \\
Statistics     & 60        & Statistical Evidence & 17     & certain       & 16         \\
Concrete       & 53        & Source credibility & 15       & analysis      & 15         \\
Photo          & 48        & Event Date        & 13        & misleading    & 12         \\
Policy         & 38        & Source Reliability & 12       & individual    & 12         \\ \hline
\end{tabular}
}
\end{table*}

\section{User query versus web retrieval human labels}
Percentage of statements for user query from human labels by category of missing information
\begin{enumerate}
    \item Speaker or person: 54.55\%
    \item Location: 76.92\%
    \item Textual context and subject specification: 47.62\%
    \item Non-textual evidence: 94.44\%
    \item Date and time period: 40.91\%
    \item Total : 59.60\%
\end{enumerate}

\section{Category classification with GPT-4}
We present results for GPT-4 category classification, using a few-shot prompt that generates a question regarding the missing information in the statement, and classifies the missing information into one of the categories we outlined. Many statements have information missing from multiple categories, and some have multiple interpretations. We therefore consider a prediction accurate when GPT-4 answer matches any one of the labels. We provide accuracy for subsets of cases when there is agreement between the labelers. We also note that not all statements have all 3 labels because of the initial disagreements on possibility of some statements (100\% of statements have 2 out of 3 labels, 69\% have all 3 labels). See Table 3 for accuracy results by category.
\begin{table*}[ht]\label{categoryclassificationtable}
\centering
\resizebox{\linewidth}{!}{%
\begin{tabular}{lcccc}
\hline
Category & Ground Truth Sample & 2/3 Majority Agreement & Full Majority Agreement \\ \hline
Speaker or person                 & 77\%                           & 68\%                                           & 76\%                                           \\ \hline
Location                 &  78\%                           & 69\%                                           & 100\%                                           \\ \hline
Textual and subject specification                 & 78\%                           & 59\%                                           & 57\%                                           \\ \hline
Non-textual evidence               & 92\%                           & 93\%                                           & 97\%                                           \\ \hline
Date and time period                & 81\%                           & 75\%                                           & 79\%                                           \\ \hline
\textbf{Total}    &  78\%                            & 71\%                                           & 80\%                                           \\ \hline
\end{tabular}
}
\caption{GPT-4 Category of missing information prediction accuracy by categories. For ground truth sample, we consider a prediction accurate if it matches any one of 3 human labels. Majority agreement cases only include statements where a 2/3 or absolute majority agreement on category exists between the labelers.}
\label{tab:my_label}
\end{table*}

\section{Prompts}\label{prompts}

\subsection{Determining Categories of Missing Information}  
Question for missing information prompt:
This is a conversation between a user and a question-answering bot. 
User: Initial Question:
Bot: To answer this question, I need to ask the following clarifying question:
Missing information keyword generation prompt: This is a conversation between a user and a question-answering bot. The bot has to limit its response to 2 words, one noun and one adjective to name the necessary detail.
User: Evaluate the truthfulness of this statement: [statement]
Bot: To evaluate the truthfulness, I need the following detail

\subsection{Category Classification and Question Generation}\label{catclassification}
Prompt:
"Classify the missing information in the statement by asking a clarifying question and assigning it to the correct category, where applicable. Use the letter associated with each category:

A: Speaker or person mentioned but not identified
B: Location mentioned but not specified
C: Textual evidence, information, or object mentioned but not specified
E: Visual or audio evidence specifically mentioned in the statement
F: Providing the date for the statement
G: Other (not related to any of the categories above)
Note: Categorize based on the most critical information missing.

Examples for reference:

A: "A nurse claims COVID is a hoax." - Which nurse are you referring to?
B: "Wages have gone up faster than inflation." - Which country are you referring to in regards to this statement?
C: "The vaccine has shown to have negative effects." - Which vaccine are you referring to?
E: "An image shows a map of Iran." - Can you provide the image you're referring to?
F: "New York has the lowest crime rate in the country." - Can you provide the date when this statement was made? (Not related to categories above)
E: "A scientist says the image shows a country's inflation grow over time." - Can you provide the image you're referring to? (In this case, the image will provide the most information.)
Given Statement: [text]

Task: Formulate the clarifying question and identify the category or categories by their respective letters, separated by a vertical bar. End your response with the category letter or letters.
\subsection{Determining When to Query User}\label{u_w}
Prompt:"Consider the following statement: [statement], and a question regarding that statement: [question]. Your task is to decide the most suitable approach for answering the question based on two options: querying the user or conducting a web search.

The user possesses the following information related to the statement, only if it is relevant for understanding the statement:

Speaker or person mentioned but not specified (pronouns)
Vague location (state, country, city, school)
Date of the statement (year, time period, day, week)
Extremely vague/unspecified subjects (unspecified type of medicine, law, program)
Any non-textual evidence is not recoverable (image, video, audio)
A web search can provide an answer to the question if it is likely to retrieve context about the statement online.

Example: 'Muslim figure: We must have pork-free menus or we will leave U.S.' - We cannot identify the figure by conducting a web search.
Example 'The US passed a law that cuts taxes for the poor in 2021' - We do not know what the law is, but we are likely to find information via web search.

Task: Determine whether a user query or web search is better suited for answering the question. Respond with a 'U' for user query, and 'W' for web retrieval. Only select 'U' for the user if it's unlikely that the statement can be retrieved from the web."

\subsection{Veracity Evaluation}
Uncertainty-enabled Prompt:
"Rate the truthfulness of the following statement: [text]."

With Context:
"The following context from a user may be provided: Question: [question], Answer: [context]."
"Please rate the statement's truthfulness on a scale from 0 to 1, where 0 signifies 'False' and 1 signifies 'True'. If uncertain or lacking context, use 0.5. Do not make assumptions or provide explanations; respond with a number."

Without Context:
"Please rate the statement's truthfulness on a scale from 0 to 1, where 0 signifies 'False' and 1 signifies 'True'. If uncertain or lacking context, use 0.5. Do not make assumptions or provide explanations; respond with a number."

Uncertainty Disabled prompt: Same as uncertainty enabled, except with "Respond with 0 or 1 to your best ability; do not provide any other responses. Do not make assumptions or provide explanations; only respond with a number." at the end.